\documentclass[10pt, a4paper]{article}
\usepackage{lrec2022} % this is the new LREC2022 Style
\usepackage{multibib}
\newcites{languageresource}{Language Resources}
\usepackage{graphicx}
\usepackage{tabularx}
\usepackage{soul}
\usepackage{titlesec}
\titleformat{\section}{\normalfont\large\bfseries\center}{\thesection.}{1em}{}
\titleformat{\subsection}{\normalfont\SmallTitleFont\bfseries\raggedright}{\thesubsection.}{1em}{}
\titleformat{\subsubsection}{\normalfont\normalsize\bfseries\raggedright}{\thesubsubsection.}{1em}{}
\renewcommand\thesection{\arabic{section}}
\renewcommand\thesubsection{\thesection.\arabic{subsection}}
\renewcommand\thesubsubsection{\thesubsection.\arabic{subsubsection}}
\usepackage{epstopdf}
\usepackage[utf8]{inputenc}

\usepackage{hyperref}
\usepackage{xstring}
\usepackage{color}

\usepackage{amssymb}
\usepackage{amsmath}

\title{HyperBox: A Supervised Approach for \\ Hypernym Discovery using Box Embeddings}

\name{Maulik Parmar, Dr. Apurva Narayan} 

\address{Independent Researcher, University of British Columbia \\
         India, British Columbia, Canada \\
         maulikres@gmail.com, apurva.narayan@ubc.ca\\}

\abstract{
Hypernymy plays a fundamental role in many AI tasks like taxonomy learning, ontology learning, etc. This has motivated the development of many automatic identification methods for extracting this relation, most of which rely on word distribution.  We present a novel model \textit{HyperBox} to learn box embeddings for hypernym discovery. Given an input term, \textit{HyperBox} retrieves its suitable hypernym from a target corpus. For this task, we use the dataset published for SemEval 2018 Shared Task on Hypernym Discovery. We compare the performance of our model on two specific domains of knowledge: medical and music. Experimentally, we show that our model outperforms existing methods on the majority of the evaluation metrics. Moreover, our model generalize well over unseen hypernymy pairs using only a small set of training data.
 \\ \newline  }

\begin{document}

\maketitleabstract

\section{Introduction}

\begin{figure}[t]
\centering
\includegraphics[scale=0.35]{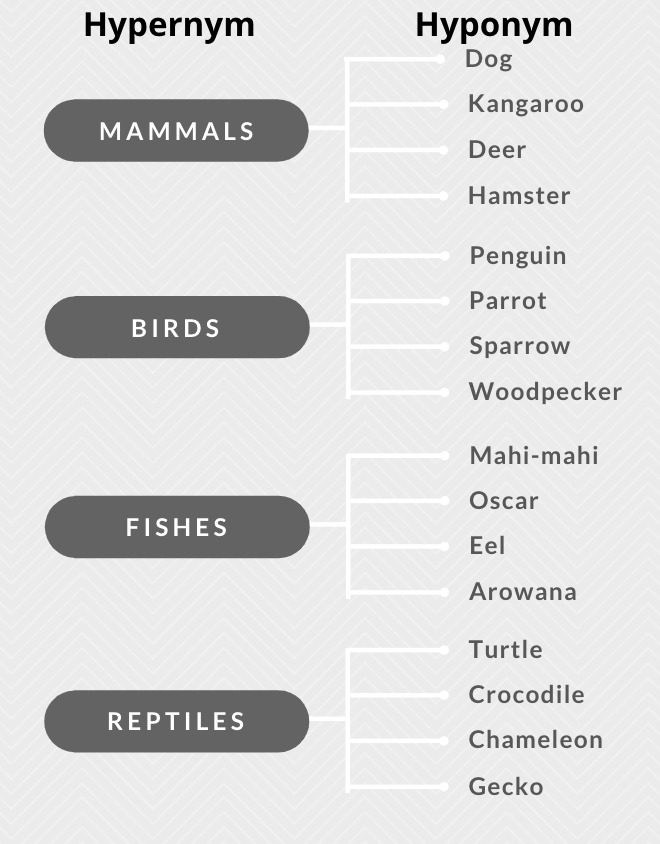}
\caption{Examples of Hypernym-Hyponym pairs}
\label{Hypernym fig}
\end{figure}

In linguistics, hypernymy is a semantic relation between a hypernym denoting a superordinate and a hyponym denoting a subordinate. Hypernymy is a major semantic relation and a vital organization principle of semantic memory \cite{MILLER&Fellbaum}. It is an asymmetric relation between a hypernym (supertype) and a hyponym (subtype), as in animal-dog and sport-tennis. Figure \ref{Hypernym fig} shows some examples of a hypernym class along with some of their hyponyms. As we can see from Figure \ref{Hypernym fig}, hypernyms are a more general class of hyponym terms. It plays a crucial role in language understanding because it enables generalization, which lies at the core of human cognition. Therefore, it has been an active area of research in NLP for decades. Automatic hypernym discovery is useful in many tasks like taxonomy creation \cite{snow-semantic,navigli}, recognizing textual entailment \cite{dagan_entail}, and text generation \cite{biran-mckeown}.
\\

In this paper, we tackle the problem of hypernym discovery \cite{espinosa-anke-etal-2016-supervised} instead of hypernym detection \cite{shwartz-etal-2017-hypernyms}. Generally, evaluation benchmarks for modeling hypernymy are such that they are reduced to a binary classification task where one tries to predict if a hypernymy relation exists between candidate pairs. Hypernym detection uses this experimental setting and tends to suffer from lexical memorization phenomena \cite{levy-etal-2015-supervised} due to the inherent modeling of the datasets by supervised systems.  Thus, to alleviate this problem \cite{espinosa-anke-etal-2016-supervised} proposed to frame the problem as Hypernym Discovery i.e given a query term and search space of domain vocabulary, discover the best candidate hypernyms of input hyponym. This reformulation not only helps in alleviating the issues discussed above but also helps to use it with other downstream applications such as semantic search, query understanding etc. Motivated by this, the organizers of SemEval Task9 published a full-fledged benchmarking dataset \citelanguageresource{camacho-collados-etal-2018-semeval} for the novel task of hypernym discovery, which covered multiple languages and knowledge domains. In this paper, we test our hypothesis on a corpus of the English language in two domains: Music and Medical.

Two families of approaches to identify and discriminate hypernyms are prominent in Hypernym Discovery. Pattern-based approaches for relation extraction have been discussed for a while in the literature and are used to discover a variety of relations including general hypernymy relation. The pattern-based approach \cite{hearst-1992-automatic,navigli-velardi-2010-learning,pavlick} to discover hypernymy was pioneered by Hearst \cite{hearst-1992-automatic} where the author defined certain lexico-syntactic patterns (e.g X such as Y) to discover hypernymy relations between pairs from corpora. Hearst introduced many such patterns in the paper for hypernym discovery. But generally, these approaches suffer from low recall as the inherent assumption is that both hypernym hyponym pairs co-occur in a pattern. This is often not the case and leads to reduced recall.

The second line of approaches uses supervised techniques and distributional models \cite{sanchez-riedel-2017-well,weeds-etal-2014-learning,santus-etal-2014-chasing} for the task of hypernym discovery. The general idea is to learn a function that takes as input the word embeddings of a query q and a candidate hypernym h and outputs the likelihood that there is a hypernymy relationship between q and h or outputs a distance in the embedding space between q and h. This decision function is learned in a supervised fashion using examples of pairs of words that are related by hypernymy and pairs that are not. 
\\\\
In this work, we consider the task of discovering hypernyms from large text corpora in a supervised way. We use the recently introduced Box Embeddings \cite{ralph_box} to discover hypernyms from a text corpus. \cite{ralph_box} proposed a spatio-translational embedding model, called BoxE that embeds entities as points, and relations as a set of hyper-rectangles (or boxes), which spatially characterize basic logical properties. Our approach is also based on Box embeddings. We show that our method \textit{HyperBox} experimentally outperforms existing methods for hypernym discovery on most of the evaluation metrics. Our contributions are as follows:

\begin{itemize}
    \item We introduce Box embeddings for hypernym discovery. To the best of our knowledge, this is the first model of its kind for hypernym discovery.
    \item Through extensive experiments on real-world datasets, we establish HyperBox’s effectiveness in discovering Hypernyms.
\end{itemize}

%Furthermore, the implementation of our model is made publicly available at %https://github.com/*******/HyperBox. 

\begin{figure*}[t]
\centering
\includegraphics[scale=0.5]{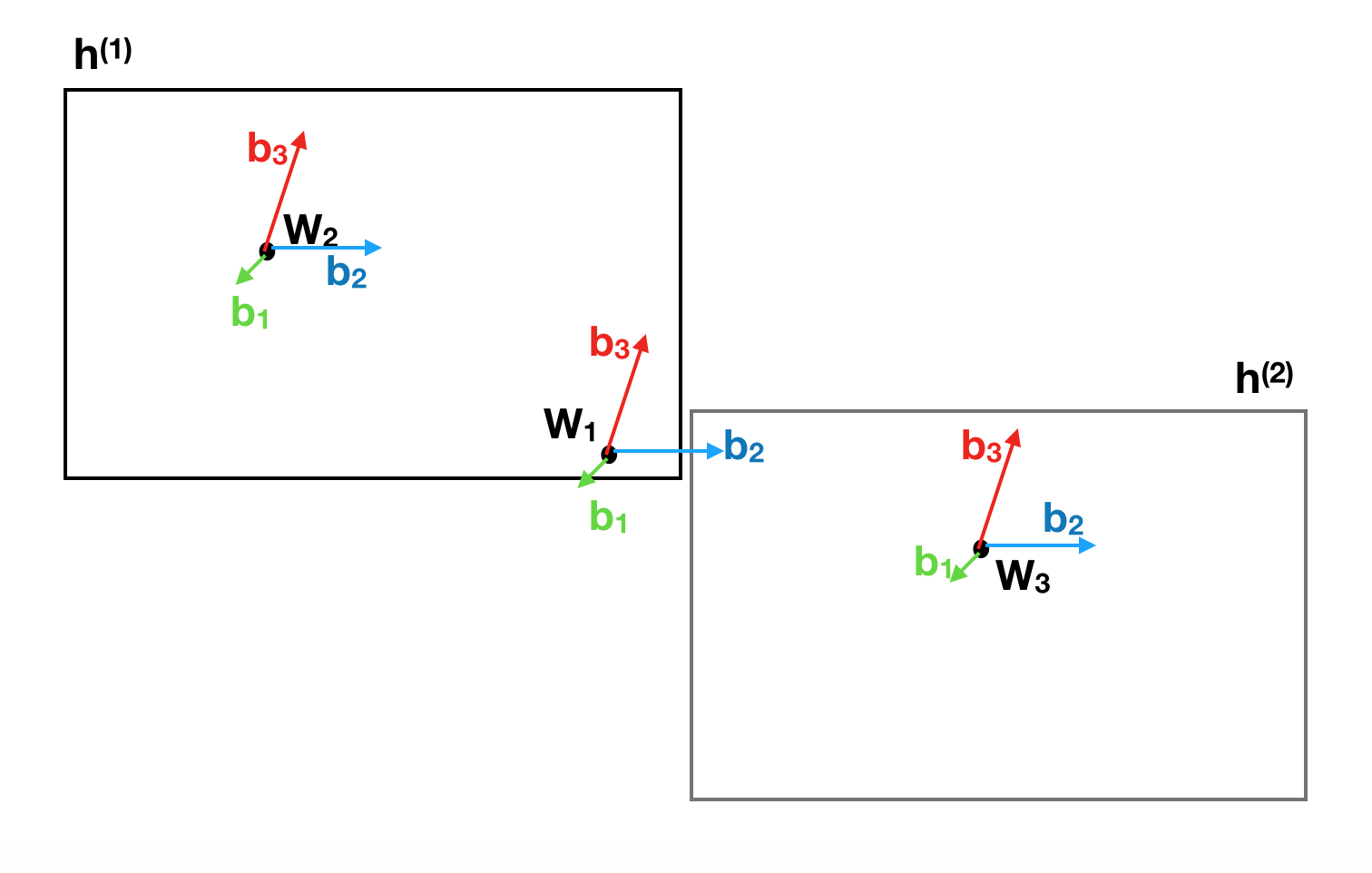}
\caption{An example \textit{HyperBox} model for three words $w_1$, $w_2$, $w_3$ in $R^2$ for hypernymy pairs $h(w_2,w_3)$, $h(w_2,w_1)$ and $h(w_1,w_3)$. The hypernymy relation is encoded by box embeddings $\mathbf{h^{(1)}}$ and $\mathbf{h^{(2)}}$. Every word $w_i$ has an embedding $\mathbf{w_i}$, and $\mathbf{b_i}$ which defines a bump on other words, as shown with distinct colors.}
\label{HyperBox fig}
\end{figure*}

\section{Related Work}
\textbf{Hypernym Detection and Discovery:}
Traditionally, discovering hypernymic relations from text corpora has been addressed using both unsupervised and supervised approaches. The pattern-based approach is a popular unsupervised approach that uses lexico-syntactic patterns to discover hypernyms from text corpora. Hearst in her paper \cite{hearst-1992-automatic} defined many such patterns for extracting hypernym relation. These high-precision patterns can also be learned automatically. However, it is well understood that the pattern-based approaches suffer significantly from missing hypernym extraction as terms must occur in exactly the right configuration to be detected.

Conversely, distributional approaches rely on a distributional representation for each observed word and are capable of discovering hypernymic relations between words even when they do not occur together explicitly in the text. Moreover, distributional approaches provide rich representations of lexical meaning. A variety of distributional methods for unsupervised hypernymy detection have been proposed \cite{weeds-weir-2003-general,lenci-benotto-2012-identifying,chang-etal-2018-distributional,weeds-etal-2004-characterising} all rely on some variation of the distributional inclusion hypothesis: If x is a semantically narrower term than y, then a significant number of salient distributional features of x is expected to be included in the feature vector of y as well. Moreover, \cite{santus-etal-2014-chasing} proposed the distributional informativeness hypothesis i.e hypernyms tend to be less informative than hyponyms, and that they occur in more general contexts than their hyponyms.

Most of the recent work on the subject is however supervised and is based on using word embeddings as input for classification or prediction \cite{fu-etal-2014-learning,espinosa-anke-etal-2016-supervised,sanchez-riedel-2017-well,baroni-etal-2012-entailment,nguyen-etal-2017-hierarchical}. \cite{shwartz-etal-2016-improving} showed that pattern-based and distributional evidence can be effectively combined within a neural architecture.
\\\\
\textbf{Embeddings:}
Our approach is based on embeddings. \cite{Yu2015} proposed a dynamic distance-margin model to learn term embeddings that capture properties of hypernymy. The model is trained on the pre-extracted taxonomic relation data and the resulting term embeddings are fed to an SVM classifier to predict hypernymy relation. However, one of the major drawbacks of this model is that they learn term pairs without considering their contexts, leading to a lack of generalization for term embeddings. Order-embeddings \cite{vendrov2016orderembeddings} represent text and images with embeddings where the ordering over individual dimensions forms a partially ordered set. 

Hyperbolic embeddings represent words in hyperbolic manifolds such as the Poincare ball and may be viewed as a continuous analogue to tree-like structures \cite{nickel_2017,nickel2018learning}. But these graph-based methods generally require supervision of hierarchical structure, and cannot learn taxonomies using only unstructured noisy data. \cite{luu-etal-2016-learning} introduced a dynamic weighting neural network to learn term embeddings that encode information about hypernymy and their contexts, considering all terms between a hyponym and its hypernym in a sentence. The proposed model is trained on a set of hypernym relations extracted from WordNet \cite{wordnet}. The embeddings are fed as features to an SVM classifier to detect hypernymy but the method still is not able to determine the directionality of a hypernym pair. \cite{vilnis-etal-2018-probabilistic,li2018smoothing} proposed construction of a novel box lattice and accompanying probability measure to capture anticorrelation and disjoint concepts. \cite{ralph_box} introduced BoxE, a Box embeddings model that embeds entities as points, and relations as a set of hyper-rectangles (or boxes), which spatially characterize basic logical properties. Our approach is also based on Box embeddings.

\section{HyperBox: Proposed Method}

In this section, we present our model for hypernym discovery, \textit{HyperBox}. The general idea is to learn a function that
takes as input the word embeddings of a query q and a candidate hypernym h and outputs the score that there is a hypernymy relationship between q and h. To discover hypernyms for a given query q (rather than classify a given pair of words), we apply this decision function to all candidate hypernyms and select the most likely candidates. In this section, we start with the description of the \textit{HyperBox} model. After this, we describe the distance function and the training objective used to train our \textit{HyperBox} model for hypernym discovery. 
\begin{table*}[t]
\centering
\begin{tabular}{|p{3cm}|p{3cm}|p{3cm}|p{3cm}|} 
 \hline
 & Term & Hypernym(s) & Source \\ %[0.5ex] 
 \hline\hline
 Medical & pulmonary embolism & pulmonary artery finding, trunk arterial embolus, embolism & SnomedCT \\
 Music & Green Day & artist, rock band, band  & MusicBrainz  \\ [1ex] 
 \hline
\end{tabular}
\caption{ Some example hyponym terms and hypernyms extracted from different sources for both domain}
\label{table:1}
\end{table*}

\subsection{Hypernym Discovery using HyperBox}

In this subsection, we describe \textit{HyperBox}, an embedding model that encodes hypernym relation as axis-aligned hyper-rectangles (or boxes) and words as points in the d-dimensional Euclidean space. 

Consider a vocabulary obtained from a corpus, which consists of a finite set $\textbf{E}$ of words. Given a word $w_i$ and word embedding dimension $m$, the model retrieves its embedding $e_i \in \mathbb R_m$ using a lookup table. These embeddings were learned beforehand on a large unlabeled text corpus. In \textit{HyperBox}, every word $w_i$ ${\Huge \in}$  $\textbf{E}$  is represented by two vectors $\mathbf{w_i}$, $\mathbf{b_i}$ $\in$ $\mathbb R^d$ in the $d$-dimensional Euclidean space, where $\mathbf{w_i}$ defines the base position of word, and $\mathbf{b_i}$ defines its translational bump, which translates all the words co-occuring in a hypernymy relation with $w_i$, from their base positions to their final embeddings by “bumping” them. We define base projection matrix $\phi_{base} \in \mathbb R^{d\times m}$ and bump projection matrix $\phi_{bump} \in \mathbb R^{d\times m}$ to obtain base position and translational bump for each word. The base position and translational bump of each word is obtained by projecting initial word embeddings using two matrix $\phi_{base} $ and $\phi_{bump}$ as follows:

\begin{equation}
\mathbf{w_i} = \phi_{base} \cdot e_i 
\end{equation}

\begin{equation}
\mathbf{b_i} = \phi_{bump} \cdot e_i 
\end{equation}

%$$\mathbf{w_i} = \phi_{base} \cdot e_i $$ 
%$$\mathbf{b_i} = \phi_{bump} \cdot e_i $$ 
%\\
The final embedding of a word $w_i$ relative to a hypernym pair $h(w_i,w_j)$ is hence given by: 

\begin{equation}
\mathbf{w{_i}{^{h(w_i,w_j)}}=(w_i+b_j)}
\end{equation}

\begin{equation}
\mathbf{w{_j}{^{h(w_i,w_j)}}=(w_j+b_i)}
\end{equation}

%$$ \mathbf{w{_i}{^{h(w_1,w_2)}}=(w_i-b_i)+\sum_{1\le j\le 2}^{}b_j }$$
%$$ \mathbf{w{_i}{^{h(w_i,w_j)}}=(w_i+b_j)}$$
%$$ \mathbf{w{_j}{^{h(w_i,w_j)}}=(w_j+b_i)}$$

Essentially, the word representation is dynamic, as every word can have a potentially different final embedding relative to a different hypernym pair. The main idea is that every word translates the base positions
of other word co-appearing in a pair, that is, for a hypernym pair $h(w_1,w_2)$, $\mathbf{b_1}$ and $\mathbf{b_2}$ translate $\mathbf{w_2}$ and $\mathbf{w_1}$ respectively, to compute their final embeddings. 

In \textit{HyperBox}, hypernym relation h is represented by 2 hyper-rectangles, i.e., boxes, $\mathbf{h^{(1)},h^{(2)}}$  $\in$ $\mathbb R^d$.  Intuitively, this representation defines two regions in $\mathbb R^d$, one for hyponym and other for hypernym, such that a fact $h(w_i,w_j)$ holds when the final embeddings of $w_i$ and $w_j$ each appear in their corresponding position box.

An example \textit{HyperBox} model is shown in Figure \ref{HyperBox fig} for d=2. Consider three words $w_1$, $w_2$, $w_3$ which are represented as a point, and a hypernymy relation is represented with two boxes $\mathbf{h^{(1)}}$ and $\mathbf{h^{(2)}}$.  Every word is translated by the bump embeddings of all other words. For example, $h(w_2,w_3)$ is a true hypernym-hyponym pair in the model, since (i) $\mathbf{w{_2}{^{h(w_2,w_3)}}=(w_2+b_3)}$ is a point in $\mathbf{h^{(1)}}$ ($\mathbf{w_2}$ appears in the head box), and (ii) $\mathbf{w{_3}{^{h(w_2,w_3)}}=(w_3+b_2)}$ is a point in $\mathbf{h^{(2)}}$ ($\mathbf{w_3}$ appears in the tail box). Similarly, $h(w_1,w_3)$ is a true hypernym-hyponym pair in the model.

\subsection{Scoring Function}

We use the scoring function introduced by \cite{ralph_box}. They define a distance function for evaluating entity positions relative to box positions such that the distance function grows slowly if a point lies inside a box (relative to the center of the box), but grows rapidly if the point is outside the box. This drive points more effectively into their target boxes and ensure they are minimally changed and can remain there once inside. Formally, let $\mathbf{u^{(i)},l^{(i)}}$ $\in $ $\mathbb R^d$ be the upper and lower boundaries of a box $\mathbf{h^{(i)}}$, respectively. Let $\mathbf{c^{(i)} = (u^{(i)}+l^{(i)})/2}$  its center and $\mathbf{\omega^{(i)} = (u^{(i)}-l^{(i)}+1)}$ its width incremented by 1. A point $\mathbf{w_i}$ is inside a box $\mathbf{h^{(i)}}$ if $\mathbf{l^{(i)} \le w_i \le u^{(i)}}$. The distance function for the given word embeddings relative to a given target box is defined as follows:

\begin{equation}
\begin{split}
&\mathbf{
dist(w{_i}{^{h(w_1,w_2)}},h^{(i)} )= }             \\
&\mathbf{
\left\{ \begin{array}{ll}
|w{_i}{^{h(w_1,w_2)}}-c^{(i)} |\oslash \omega^{(i)} & if w{_i} \in h^{(i)} \\
|w{_i}{^{h(w_1,w_2)}}-c^{(i)} |\circ \omega^{(i)}-\kappa & otherwise
\end{array}
\right.
}
\end{split}
\end{equation}

%$$\mathbf{
%dist(w{_i}{^{h(w_1,w_2)}},h^{(i)} )= }$$
%$$\mathbf{
%\left\{ \begin{array}{ll}
%|w{_i}{^{h(w_1,w_2)}}-c^{(i)} |\oslash \omega^{(i)} & if w{_i} \in h^{(i)} %\\
%|w{_i}{^{h(w_1,w_2)}}-c^{(i)} |\circ \omega^{(i)}-\kappa & otherwise
%\end{array}
%\right.
%}
%$$
%\\
where $\mathbf{\kappa = 0.5 \circ (\omega^{(i)}-1) \circ (\omega^{(i)}-\omega^{(i)^{\circ-1}})}$, is a width-dependent factor, $\circ$ is element wise multiplication, $\circ -1$ is element-wise inversion and $\oslash$ is element wise division.

The distance function, $\mathbf{dist}$ factors in the size of the target box in its
computation for both the above cases. In the first case, when the point is in its target box, distance inversely correlates with box size, to maintain low distance inside large boxes and provide a gradient to keep points
inside. In the second case, box size linearly correlates with
distance, to penalize points outside larger boxes more severely.
Moreover, $\kappa$ is subtracted to preserve function continuity. More details about this distance function can be found out in \cite{ralph_box}. Finally, the scoring function is defined as the sum of the L-$2$ norms of $\mathbf{dist}$ between both hyponym-hypernym pair and their respective boxes, i.e.:

\begin{equation}
score(h(w_1,w_2)) = \sum_{i=1}^{2} ||\textbf{dist}( \textbf {w}{_i}{^{h(w_1,w_2)}},\textbf{h}^{(i)} )||_2
\end{equation}

%$$
%score(h(w_1,w_2)) = \sum_{i=1}^{2} ||dist( \textbf %{w}{_i}{^{h(w_1,w_2)}},\textbf{h}^{(i)} )||_2
%$$

\subsection{Training Objective}
Our next goal is to learn base and bump embeddings for each word as well as projection matrix and box embedding for both boxes. Given a training set of queries and their output, we optimize a negative sampling loss \cite{mikolov2013efficient} to effectively optimize a distance-based model \cite{sun2018rotate}:

\begin{equation}
L=-log\sigma(\gamma - dist(v;q))-\sum_{i=1}^{k}(1/k)log\sigma(dist(\acute{v}_i;q)-\gamma) \end{equation}
\\
%$$L=-log\sigma(\gamma - %dist(v;q))-\sum_{i=1}^{k}(1/k)log\sigma(dist(\acute{v}_i;q)-\gamma) $$

where $\gamma$ represents a fixed scalar margin, $v \in [[q]]$ is a positive entity (i.e., answer to the query q), and $\acute{v}_i \notin [[q]]$ is the i-th negative entity (non-answer to the query q) and k is the number of negative samples.

\begin{table}[h]
\centering
\begin{tabular}{||c c c||} 
 \hline
  Split & Music & Medical \\ [0.5ex] 
 \hline\hline
 Trial/Validation & 15 & 15 \\ 
 Training & 500 & 500 \\
 Test & 500 & 500  \\ [1ex] 
 \hline
\end{tabular}
\caption{ Number of terms (hyponyms) for each dataset in trial, training and test sets.}
\label{table:2}
\end{table}

\begin{table*}[t]
\centering
\begin{tabular}{|p{9cm}|p{1cm}|p{1cm}|p{1cm}|} 
 \hline
 Model & MRR & MAP & P@5 \\ [0.5ex] 
 \hline\hline
 Hypernyms under Siege \cite{shwartz-etal-2017-hypernyms} & 5.01 & 1.95 & 2.15\\
 Adapt \cite{adapt}  &	7.46 &	2.63 &	2.64\\
 SJTU \cite{sjtu}	& 9.15	& 4.71	& 4.91\\
 vTE \cite{espinosa-anke-etal-2016-supervised}	& 39.36	& 12.99	& 12.41\\
 300-sparsans \cite{300-sparsans}	& 46.43	& 29.54	& 28.86\\
 CRIM Supervised \cite{bernier-colborne-barriere-2018-crim}	& 57.34	& 39.95	& 43.00\\
 HyperBox(Our)	& \textbf{58.15} & \textbf{41.39} & \textbf{43.13}\\ [1ex] 
 \hline
\end{tabular}
\caption{ Results on the Music dataset}
\label{table:3}
\end{table*}

\section{Experiments}
\subsection{Datasets}

We use the SemEval-2018 Task9-Hypernym Discovery dataset \cite{camacho-collados-etal-2018-semeval} for our experiments. We use two datasets in the English language corresponding to two specific domains of medical and music. Some example input-output pairs (i.e. hyponym terms and corresponding hypernym lists) are shown in Table \ref{table:1} for both datasets. Table \ref{table:1} also reports the sources of hypernymy information beside each pair, which vary depending on the dataset.

Statistics of the two datasets used in the experiments of this paper are summarized in Table \ref{table:2}. The dataset was split equally in the training and testing set, while the trial data provided fewer examples and is to be used as a validation set. It should be noted that each term may be associated with one or (in most cases) more than one hypernym. Therefore, the effective number of hyponym-hypernym pairs for both datasets would be high. For example, the number of hyponym-hypernym pairs in the test gold standard is 4,116 for the medical dataset and 5,233 for the music dataset.

\subsection{Evaluation Metrics}

To compare the performance of our model with existing models on the SemEval-2018 Task9 dataset \cite{camacho-collados-etal-2018-semeval}, we also use the same set of metrics provided by the organizer of SemEval-2018 Task9. The hypernym discovery task is evaluated as a soft ranking problem. Models were evaluated over the top 15 (at most) hypernyms retrieved for each input term, and their performance was assessed through Information Retrieval metrics.

\begin{itemize}
    \item \textbf{Mean Reciprocal Rank (MRR) }: For a single query, reciprocal rank is 1/rank where rank is the position of the first correct result in a ranked list of outcomes. For multiple queries Q, the MRR is defined as follows:
    
    \begin{equation}
    MRR = \frac{1}{|Q|} \sum_{i=1}^{|Q|} 1/rank_i
    \end{equation}

    %$$MRR = \frac{1}{|Q|} \sum_{i=1}^{|Q|} 1/rank_i$$
    \item \textbf{Mean Average Precision (MAP)}: MAP is a widely used metric to measure the performance of models in information retrieval. It is defined as:
    
    \begin{equation}
    MAP= \frac{1}{|Q|} \sum_{q \in Q}^{}AP(q)
    \end{equation}

    %$$MAP= \frac{1}{|Q|} \sum_{q \in Q}^{}AP(q)$$
    where Q is the number of queries or experimental runs, AP(·) refers to average precision, i.e. an average of the correctness of each individual obtained hypernym from the search space.
   
    \item \textbf{Precision@k (P@k)}: In addition to MRR and MAP, P@k is used which is defined as the number of correctly retrieved hypernyms at various thresholds (k=1,3,5,15, etc).
    
    \begin{equation}
    P@k= \frac{true positives @k}{(true positives @k) + (false positives @k)}
    \end{equation}

    %$$P@k= \frac{true positives @k}{(true positives @k) + (false positives %@k)}$$
    
\end{itemize}

\subsection{Experimental Setup}

Training for the HyperBox model was conducted on an Intel Xeon CPU with 16 cores and 224 GB RAM.
We run \textit{HyperBox} on both the medical and music dataset. Initially, we train our word embeddings on a given raw corpus with an embedding dimension of 300. We use the same raw corpora that were provided by the organizers for the SemEval2018 Task9. For the medical dataset, a combination of abstracts and research papers provided by the MEDLINE (Medical Literature Analysis and Retrieval System) repository, which contains academic documents such as scientific publications and paper abstracts, is used. For the music domain, the raw corpus is a concatenation of several music-specific corpora, i.e., music biographies from Last.fm contained in ELMD 2.0 \cite{oramas-etal-2016-elmd}, the music branch from Wikipedia, and a corpus of album customer reviews from Amazon \cite{oramas2017multilabel}. The raw corpora along with training and test data can be downloaded from \textbf{https://competitions.codalab.org/competitions/17119}.

\textit{HyperBox} is trained using the Adam optimizer, to optimize negative sampling loss. Hyperparameter tuning was conducted over its learning rate, loss margin $\gamma$, dimensionality d, and the number of negative examples. We only report final hyperparameter values after hyperparameter tuning. We used an embedding size of 300 for both base and bump embeddings, and also for hypernym boxes (center and offset). Based on the performance on the validation set, we use a learning rate of 0.001, no of negative samples equal to 100, and $\gamma$ (margin in negative sampling loss) equal to 2.

\begin{table*}[t]
\centering
\begin{tabular}{|p{9cm}|p{1cm}|p{1cm}|p{1cm}|} 
 \hline
 Model & MRR & MAP & P@5 \\ [0.5ex] 
 \hline\hline
 Hypernyms under Siege \cite{shwartz-etal-2017-hypernyms} & 2.10 & 0.91 & 1.08\\
 Adapt \cite{adapt}	& 20.56	& 8.13	& 8.32\\
 SJTU \cite{sjtu}	& 25.95	& 11.69	& 11.69\\
 vTE \cite{espinosa-anke-etal-2016-supervised}	& 41.07	& 18.84	& 20.71\\
 300-sparsans \cite{300-sparsans}	& 40.60	& 20.75	& 21.43\\
 CRIM supervised \cite{bernier-colborne-barriere-2018-crim}	& 37.63	& \textbf{28.51}	& 25.63\\
 HyperBox(Our)	& \textbf{43.71} & 27.79 & \textbf{30.22}\\ [1ex] 
 \hline
\end{tabular}
\caption{ Results on the Medical dataset}
\label{table:4}
\end{table*}

\section{Results}
A summary of the results is provided in Tables \ref{table:3} and \ref{table:4}. We compare the performance of our model with the benchmark methods and models of the team participating in the SemEval 2018 Task9. It is worth noting that we don't use hyperbolic embeddings for comparison. Hyperbolic embeddings \cite{nickel_2017,nickel2018learning} have been shown to perform well in learning the hierarchical structure as observed in trees. But, to use hyperbolic embeddings we need a graph-structured dataset instead of the raw corpus. \cite{le-etal-2019-inferring} use Hearst Graphs to create such graph-like structure from the raw corpus. But for the given dataset, such a graph will be very sparse with many disconnected components. So in this work, we skip comparison with models using hyperbolic embeddings.

The Adapt team \cite{adapt} uses skip-gram word embeddings for hypernym discovery. They use the traditional word2vec similarity function to discover hypernym from a raw corpus. The SJTU team \cite{sjtu} uses neural term embeddings for hypernym discovery. They use different neural networks like LSTM, CNN, GRU to learn term embeddings to discover hypernym from a raw corpus. The vTe team uses a supervised distributional framework for hypernym discovery which operates at the sense level, by exploiting semantic regularities between hyponyms and hypernyms in embeddings spaces and integrating a domain clustering algorithm \cite{espinosa-anke-etal-2016-supervised}. The 300-sparsans \cite{300-sparsans} team uses a system based on sparse coding and a formal concept hierarchy obtained from word embeddings. The CRIM team \cite{bernier-colborne-barriere-2018-crim} uses a hybrid approach by combining methods based on unsupervised Hearst patterns and supervised projection learning. We use the supervised model of CRIM \cite{bernier-colborne-barriere-2018-crim} for a fair comparison with all the existing models. Most of these models use symmetric similarity functions. As a result of this, even if we interchange hyponym-hypernym pairs their symmetric similarity function will output the same score which is undesirable. Unlike this, \textit{HyperBox} doesn't face such a problem because it uses a Box structure and order.

We can see from Tables \ref{table:3} and \ref{table:4} that our method \textit{HyperBox} outperforms all existing benchmark models on most of the metrics. The SOTA results are bolded in the table. This is because our \textit{HyperBox} model is able to learn the anti-symmetric and hierarchical relation "hypernymy" very well. \cite{ralph_box} showed that the Box embedding model is fully expressive, and is capable of learning symmetry, anti-symmetry, inversion, composition, hierarchy, intersection, and mutual exclusion.

\section{Conclusion}

\textit{HyperBox} provides an effective way for solving the problem of Hypernym discovery. Unlike the Hypernym detection task which reduces to a binary classification task, hypernym discovery focuses on retrieving hypernyms from a large text corpus. \textit{HyperBox} encodes words as points and hypernym relation as axis-aligned hyper-rectangles (or boxes) in the d-dimensional euclidean space. Moreover, Box embeddings have been shown to learn antisymmetric and hierarchical relations very well due to their distinctive box structure.

As part of future work, we hope to combine \textit{HyperBox} with existing unsupervised approaches like Hearst patterns to form a hybrid approach to solve the problem of hypernym discovery.

% \nocite{*}
\section{Bibliographical References}\label{reference}
%\label{main:ref}

\bibliographystyle{lrec2022-bib}
\bibliography{lrec2022-example}

\section{Language Resource References}
\label{lr:ref}
\bibliographystylelanguageresource{lrec2022-bib}
\bibliographylanguageresource{languageresource}

\end{document}